\documentclass{article}

\usepackage{arxiv}

\usepackage[]{algorithm}
\usepackage{algpseudocode}
\algrenewcomment[1]{\(\triangleright\) #1}
\algnewcommand{\LineComment}[1]{\State \(\triangleright\) #1}

\usepackage{acronym}
\usepackage{ragged2e}
\usepackage{amsthm}
\usepackage{amsmath}
\usepackage{cases}
\usepackage{caption}
\usepackage{graphicx}
\usepackage{subcaption}
\usepackage{amsfonts}
\usepackage{scalerel}
\usepackage{xcolor}
\usepackage{authblk}
\usepackage{listings}
\usepackage{hyperref}
\usepackage{url}
\usepackage[inline]{enumitem}
\acrodef{asp}[ASP]{Answer Set}
\acrodef{las}[LAS]{Learning from Answer Sets}
\acrodef{ilp}[ILP]{Inductive Logic Programming}
\acrodef{wcdpi}[WCDPI]{weighted context-dependant partial interpretation}

\theoremstyle{definition}
\newtheorem{mydef}{Definition}
\newtheorem{myexample}{Example}

\newcommand{\pluseq}{\mathrel{+}=}
\newcommand{\asp}[1]{\mbox{$\mathtt{#1}$}}

\title{NSL: Hybrid Interpretable Learning From Noisy Raw Data\thanks{This article has been updated, please refer to \href{https://arxiv.org/abs/2106.13103}{FF-NSL: Feed-Forward Neural-Symbolic Learner}~\cite{cunnington2021ffnsl}}}

\author[1,2]{Daniel Cunnington}
\author[2]{Alessandra Russo}
\author[2]{Mark Law}
\author[2]{Jorge Lobo}
\author[3]{Lance Kaplan}
\affil[1]{IBM Research Europe, Winchester, UK.\\Contact: \texttt{dancunnington@uk.ibm.com}}
\affil[2]{Imperial College London, London, UK}
\affil[3]{Army Research Laboratory, Adelphi, MD, USA}

\begin{document}
\maketitle

\begin{abstract}
Inductive Logic Programming (ILP) systems learn generalised, interpretable rules in a data-efficient manner utilising existing background knowledge. However, current ILP systems require training examples to be specified in a structured logical format. Neural networks learn from unstructured data, although their learned models may be difficult to interpret and are vulnerable to data perturbations at run-time. This paper introduces a hybrid neural-symbolic learning framework, called \textit{NSL}, that learns interpretable rules from labelled unstructured data. NSL combines pre-trained neural networks for feature extraction with FastLAS, a state-of-the-art ILP system for rule learning under the answer set semantics. 
Features extracted by the neural components define the structured context of labelled examples and the confidence of the neural predictions determines the level of noise of the examples. Using the scoring function of FastLAS, NSL searches for short, interpretable rules that generalise over such noisy examples. We evaluate our framework on propositional and first-order classification tasks using the MNIST dataset as raw data. Specifically, we demonstrate that NSL is able to learn robust rules from perturbed MNIST data and achieve comparable or superior accuracy when compared to neural network and random forest baselines whilst being more general and interpretable.
\end{abstract}


\section{Introduction}
\ac{ilp} systems learn (a set of) logical rules, that together with some background knowledge, explain a set of examples. \ac{ilp} systems are often praised for their data efficiency~\cite{cropperturning,lin2014bias} and the interpretable nature of their learned rules~\cite{muggleton2018}. Recent state-of-the-art \ac{ilp} systems have also shown to be noise-tolerant and capable of learning complex knowledge from mislabelled data in
an effective manner~\cite{Law2018thesis,law2020fastlas}. A common characteristic of all state-of-the-art \ac{ilp} systems is the fact that examples are required to be specified in a structured, logical format. This may hinder their applicability to real-world domains, leaving, as one of their main outstanding challenges, the ability to learn from unstructured data where contextual information may be noisy or perturbed.

Differentiable learning systems, such as deep neural networks~\cite{Goodfellow-et-al-2016}, have demonstrated powerful function approximation on a wide variety of tasks, solving classification problems directly from unstructured data. However, large amounts of training examples are required, learned models are difficult to interpret~\cite{8631448} and may be vulnerable to distributional shift, where simple data perturbations are observed at run-time~\cite{sensoy2018evidential}. In this case, neural network predictions may be incorrect and the focus of this paper is learning robust rules in the presence of incorrect neural network predictions, taking into account the confidence score of such predictions.

To achieve this, we introduce a hybrid Neural-Symbolic Learning framework, called \textit{NSL}, that aims to combine the advantages of both \ac{ilp} and differentiable learning systems. NSL aims to learn a set of interpretable rules, that together with a (possibly empty) background knowledge, explain a given set of labelled unstructured data. It does so, by using a pre-trained neural network for extracting features from the unstructured data, and automatically generating structured context-dependent examples that are used by a state-of-the-art \ac{ilp} system, FastLAS~\cite{law2020fastlas} to learn rules that explain the given labelled unstructured data. The neural network predictions form the \textit{context} of FastLAS's labelled examples and the confidence of the neural network predictions are aggregated to define a notion of penalty for these examples. FastLAS uses this notion of penalty, together with the length of rules, to define a cost over possible solutions. Rules that do not cover a context-dependent example pay the penalty of the example. Optimal solutions are computed as the set of rules with minimal cost. This enables NSL to cater for neural predictions with low confidence due to perturbations in the unstructured data and still learn rules that generalise (i.e. shorter rules pay a lower cost) and maximise coverage of labelled unstructured data with the highest aggregated confidence of neural predictions.

The NSL framework is evaluated on two classification tasks, a propositional animal classification task and a first-order valid Sudoku board classification task, where numerical features are extracted from MNIST digit images~\cite{lecun1998gradient}. The neural network component is pre-trained on \textit{non-perturbed} MNIST digits. In each task, the NSL framework is trained by perturbing the MNIST digits -- rotating each digit image 90$^{\circ}$ clockwise and rules are learned that define animal class and the notion of valid Sudoku boards respectively. The learned rules are evaluated on structured ground truth test data and unstructured perturbed test data to, respectively (i) validate the accuracy of the learned rules with respect to a clean dataset (even though the rules were learned in the presence of perturbation at training time), and (ii) evaluate the robustness of the learned rules when applied to unseen perturbed unstructured data. In the first case and in both learning tasks, NSL is able to learn robust rules in the presence of training data perturbations, achieving comparable or superior accuracy than neural network and random forest baselines. Also, NSL learns rules that are more general and more interpretable. In the second case, when perturbations occur at run-time, NSL is able to maintain robust classification performance up to $\sim$40\% perturbations, outperforming the neural network and random forest baselines. Finally, in the Sudoku classification task, we also demonstrate NSL's data efficiency compared to the neural network baseline, as the neural network requires 12.5X the number of examples required by NSL to achieve comparable performance. 


The paper is structured as follows. In the next section we review relevant background material. We then introduce our NSL framework for learning rules from unstructured data, and discuss its algorithms. We then follow with results of our extensive evaluation. Finally we conclude with a discussion of related work.
\section{Background}\label{sec:background}
This section briefly introduces notations and terminologies used throughout the paper. An \ac{ilp} learning task aims to find a set of rules, called a hypothesis, that explains a set of labelled examples~\cite{Muggleton1991}. Different approaches have been proposed in the literature~\cite{muggleton1994inductive} and in this paper we focus on the~\ac{las} approach~\cite{LawRB19}, as the recently proposed ILASP systems have been shown to be robust to noisy examples~\cite{law2018inductive} and scalable to large hypothesis spaces~\cite{law2020fastlas}. An \ac{asp} program formalises a given problem in a logical form so that solutions to the program, called {\em answer sets}, provide solutions to the original problem. Formally, an \ac{asp} program is a set of rules of the form $\asp{h}:\!-\; \asp{b_{1}}\ldots \asp{b_{n}}, \asp{not}\;\asp{c_{1}},\ldots\asp{not}\;\asp{c_{m}}$ where $\asp{h}$, $\asp{b_{i}}$ and $\asp{c_{j}}$ are atoms; $\asp{h}$ is the {\em head} of the rule, and $\asp{b_{1}}\ldots\asp{not}\;\asp{c_{m}}$ is the {\em body} of the rule, formed by a conjunction of positive literals ($\asp{b_{1}}\ldots\asp{b_n}$) and negative literals $(\asp{not}\;\asp{c_{1}}\ldots\asp{not}\;\asp{c_{m}})$ where $\asp{not}$ is negation as failure. Given an ASP program $P$, the Herbrand Base of $P$ (HB($P$)) is the set of ground atoms constructed using the predicates and constants that appear in $P$. An interpretation $I$ is a subset of HB($P$). Given an interpretation $I$, the reduct of the grounding of an ASP program $P$, denoted as $P^{I}$ (see~\cite{Gelfond88} for definition and algorithm) is a grounded program with no negation as failure. $I$ is an answer set of $P$ if and only if it is the minimal model of the reduct program $P^{I}$. We denote the set of answer sets of a program $P$ with $AS(P)$.

In the \ac{las} framework, the objective is to learn \ac{asp} programs from examples that are {\em partial interpretations}. A \textit{partial interpretation} $e_{pi}$ is a pair of sets of ground atoms $\left \langle e^{inc}, e^{exc} \right \rangle$, called {\em inclusion} and {\em exclusion} sets respectively~\cite{law2020fastlas, law2018inductive}. An interpretation $I$ \textit{extends} $e$ iff $e^{inc} \subseteq I$ and $e^{exc} \cap I = \emptyset$. In ILASP~\cite{law2018inductive}, an example can have an associated context and penalty value. In this case, the example is called a \ac{wcdpi}. This is defined as a tuple $e = \left \langle e_{id}, e_{pen}, e_{pi}, e_{ctx} \right \rangle$ where $e_{id}$ is a unique identifier for $e$, $e_{pen}$ is either a positive integer or $\infty$, called a {\em penalty}, $e_{pi}$ is a partial interpretation and $e_{ctx}$ is an \ac{asp} program relative to the example, called {\em context}. A \ac{wcdpi} $e$ is {\em accepted} by a program $P$ if and only if there is an answer set of $P \cup e_{ctx}$ that extends $e_{pi}$. A task that learns ASP programs from \acp{wcdpi} is called a context-dependent~\ac{las} task and denoted as $ILP^{context}_{LAS}$. Such a task is defined as a tuple $\langle B,S_M,E \rangle$, where $B$ is background knowledge, expressed as an \ac{asp} program, $S_M$ is the hypothesis space, defined by a language bias $M$\footnote{For a detailed definition of a language bias see~\cite{Law2018thesis}.}, and $E$ is a set of  \acp{wcdpi}. The hypothesis space defines the set of rules that can be used to construct a solution to the task. A hypothesis $H\subset S_{M}$ is a solution to a given $ILP^{context}_{LAS}$ task if and only if for every $e\in E$, $H$ covers $e$, that is $B\cup e_{ctx}\cup H$ accepts $e$. 

FastLAS is a system for solving a specific class of $ILP^{context}_{LAS}$ learning tasks, where for all $e \in E$, $\left | AS(B \cup e_{ctx})  \right | = 1$ and no predicate in the head of a rule in $S_{M}$ may occur in the body of any rule in $S_{M}$ or in $B$~\cite{law2020fastlas}. Such tasks are referred to in the literature as Observational Predicate Learning (OPL) tasks. In this paper we will denote them as $ILP^{OPL}_{LAS}$, and the set of possible solutions of a given $ILP^{OPL}_{LAS}$ task $LT$, as $ILP^{OPL}_{LAS}(LT)$.  Unless otherwise specified, we will assume that our $ILP^{OPL}_{LAS}$ tasks have \acp{wcdpi} as examples. 

The FastLAS system uses a {\em scoring function} to compute optimal solutions of a given learning task $LT = \langle B, S_{M}, E\rangle$. This is a function $\mathcal{S} : Programs \times T_{OPL} \rightarrow \mathbb{R}_{\geq0}$ where $Programs$ is the set of all \ac{asp} programs and $T_{OPL}$ is the set of all OPL tasks defined above (see~\cite{law2020fastlas} for a formal definition). A hypothesis $H \in Programs$ is said to be an {\em optimal solution} of a task $LT \in T_{OPL}$ with respect to a scoring function $\mathcal{S}$ iff $H \in ILP^{OPL}_{LAS}(LT)$ and there is no $H^{\prime} \in ILP^{OPL}_{LAS}(LT)$ such that $\mathcal{S}(H^{\prime},LT) < \mathcal{S}(H,LT)$. Scoring functions can be domain-specific and can be given as input to FastLAS prior to solving a learning task. FastLAS is capable of supporting any scoring function that is {\em decompositional}, that is for each $P \in Programs$ and $LT \in T_{OPL}$:
\[\mathcal{S}(P, LT) = \sum\limits_{r \in P}\mathcal{S}_{rule}(r, LT)\]
where $\mathcal{S}_{rule}: Rules \times T_{OPL} \rightarrow \mathbb{R}_{\geq0}$ is a function that scores individual rules of a solution $H$. We refer to $\mathcal{S}_{rule}$ as the decomposition of $\mathcal{S}$.

\section{NSL Framework}
\label{framework}
\begin{figure*}[t!]
  \centering
  \includegraphics[keepaspectratio, width=1\textwidth]{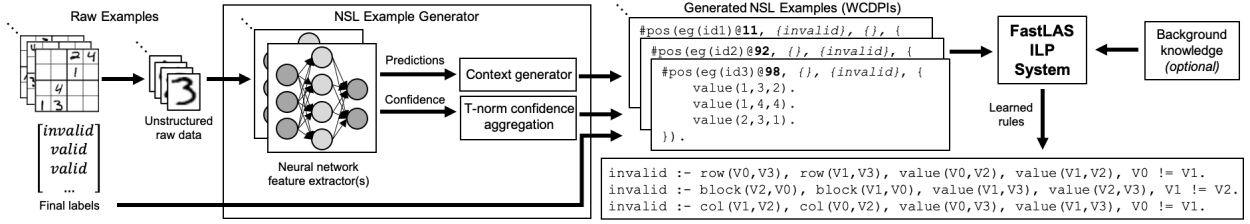}
  \caption{NSL framework with an example Sudoku board classification task }
  \label{fig:nsl_arch}
\end{figure*}

This section introduces our Neural-Symbolic Learning (NSL) framework for learning rules from unstructured data. The framework uses two main computational components: a neural network, for extracting features from unstructured data, and FastLAS for learning rules from WCDPI examples. The bridge between these two components is realised through the {\em NSL example generator}, which converts unstructured labelled data into structured examples by generating a WCDPI example for FastLAS, from each set of neural predictions and a given label. A high level overview of the NSL framework, applied to a Sudoku classification task, is shown in Figure~\ref{fig:nsl_arch}. The aim of the Sudoku task is to learn the classification of Sudoku boards as \textit{valid} or \textit{invalid}. The input is a set of images of digits in a Sudoku board with a valid/invalid label. The NSL framework returns as output an optimal set of rules that define the notion of an invalid Sudoku board. These rules can then be used to classify unseen Sudoku boards (given as sequences of digit images) as valid or invalid (see evaluation section). In what follows we present in detail the {\em NSL example generator} and the scoring function used by FastLAS as {\em optimisation criteria}, based on aggregated confidence of the neural network feature predictions. We then formalise the NSL learning task. 

\subsection{NSL Example Generator}
The neural component  of our NSL framework is a set of feature extractors $\mathcal{N}$, each pre-trained to extract a specific feature from a given piece of unstructured raw data. 
Let $d \in D$ be a piece of unstructured raw data (e.g. an image) and $t$ be a string representing a specific feature of the unstructured raw data (e.g. a \textit{digit}). A sample input $x$ to the neural component is a pair $\langle d, t\rangle$ for which a feature extractor $n_{t}$ exists in $\mathcal{N}$. For each sample input $x=\langle d,t\rangle$, the feature extractor $n_{t} \in \mathcal{N}$ outputs a $k$-dimensional probability vector $\rho_{d}$ over the $k$-class feature $t$ of $d$, i.e. $\rho_{d} = n_{t}(d)$. Within the scope of this paper, we assume that a specific feature extractor $n_{t}$ can only extract feature $t$ from a piece of unstructured raw data $d$ of a given sample $x = \langle d, t\rangle$. Let $X$ be a set of such samples.

We assume that each $n_{t}\in\mathcal{N}$ is a \emph{pre-trained} neural network with fixed weights performing a classification task on input $d$. Each feature extractor $n_{t}$ generates a prediction $y_{pred}^{t} = argmax(\rho_{d})$ with a level of confidence $y_{conf}^{t} = max(\rho_{d})$. These represent, respectively, the prediction of feature $t$ and associated confidence estimation output by the neural network $n_{t}$ for a given input $d$.




For each piece of unstructured raw data $d_{i} \in D$, the confidence score $y_{conf}^{t_{i}}$ 
of the corresponding feature extractor in $\mathcal{N}$ is used by the NSL example generator to define the penalty of a WCDPI example constructed from $D$. This is achieved by means of a {\em general aggregation} function that combines a set of $y_{conf}^{t_{i}}$ confidence estimations 
into a single confidence score value. So, given a set of unstructured raw data $D$ and an associated set of samples $X_{D} = \{\langle d_{i}, t_{i}\rangle\}$, the {\em overall confidence estimation} for $D$ is an aggregation of the confidence estimation of each sample. This aggregated estimation, denoted as $y^D_{conf}$, constitutes a single confidence estimation value for a set of features extracted from unstructured raw data $D$. 

The features extracted from unstructured raw data form the atomic building blocks for the \textit{context} of the generated WCDPI examples, $e_{ctx}$. In FastLAS, contexts can be any \ac{asp} program which imposes additional example-based constraints that are taken into account during the learning task. In this paper, we consider a context to be a conjunction of atoms. 
To define the overall confidence estimation of a context $e_{ctx}$ for an example $e$, NSL uses an aggregation function $\mathcal{A}$ defined in terms of triangular norms, (or \textit{t}-norms)~\cite{klement2013triangular}. Given that a context of a generated example is a conjunction of atoms, NSL uses the Gödel \textit{t}-norm ($\mathcal{T}_g(x,y)=min(x,y)$) (\cite{metcalfe2008proof} and~\cite{rbac2010fuzzy}) to recursively define the aggregation function $\mathcal{A}$. This is formally defined below. The aggregated confidence estimation $y^D_{conf}$ defines the penalty value, $e_{pen}$, of a \ac{wcdpi} example $e$, whose context $e_{ctx}$ has been constructed from the predictions $y_{pred}^{t}$ of a set of neural network feature extractors.


\begin{mydef}[General aggregation function]
\label{def:aggregation_function}
Let $\mathcal{T}_{g}:[0,1]^{2}\rightarrow [0,1]$ be the Gödel \textit{t}-norm function defined as $\mathcal{T}_g(x,y)=min(x,y)$. Let $\mathcal{A}_{z}: [0,1]^z \rightarrow [0,1]$, where $z\geq 2$, be an aggregation function, recursively defined as follows\footnote{Due to the associativity property of $T_{g{}}$ the recursive case is the same for all $1\leq j\leq z$.}: 
\begin{equation*}
  \begin{split}
    \mathcal{A}_{2}(x_1,x_2) &= \mathcal{T}_g(x_1,x_2)\\
    \mathcal{A}_{z}(x_1,\ldots,x_z) &= \mathcal{T}_g(A_{j}(x_1,\ldots,x_j), A_{z-j}(x_{j+1},\ldots, x_{z}))\\   
  \end{split}
\end{equation*}

A {\em general aggregation function} 
$\mathcal{A}: [0,1]^z \rightarrow \mathbb{R}$ is then given by $\mathcal{A}=\lfloor \lambda \mathcal{A}_{z} \rfloor$ for $z\geq 2$, where $\lambda$ is a parameter used alongside the floor function to convert the aggregated confidence score $\mathcal{A}_{z}([0,1]^z) \in [0,1]$ to an integer value. 
\end{mydef}

\begin{mydef}[Context Generator]\label{def:build_context_function}
Let $F$ be the set of all possible sets of 
tuples $\langle y_{pred}, t,\alpha \rangle$ indicating a prediction $y_{pred}$ for feature $t$. $\alpha$ can be used to encode additional information related to a particular feature prediction, such as the coordinates of a digit in a Sudoku board classification task. A context generator is a \textit{function} $\mathcal{G}: F \rightarrow C$, where $C$ is the set of all possible contexts for a \ac{wcdpi}. Given an $f_{i} \in F$, 
$\mathcal{G}(f_{i}) = \{t(\alpha,y_{pred}). \mid 
\langle y_{pred}, t, \alpha \rangle\in f_{i}\}$.\footnote{Note that a more complex function $\mathcal{G}$ could be engineered, but this would require additional supervision.}
\end{mydef} Note that $\mathcal{G}(f_{i})$ gives the context $e_{ctx}$ of an example for the FastLAS component based on the neural feature predictions.

\begin{myexample}\label{ex:ctx_gen}
Let us assume a learning task with a set of unstructured raw data $D = \left \{\text{\scalerel*{\includegraphics{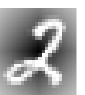}}{)}}, \text{\scalerel*{\includegraphics{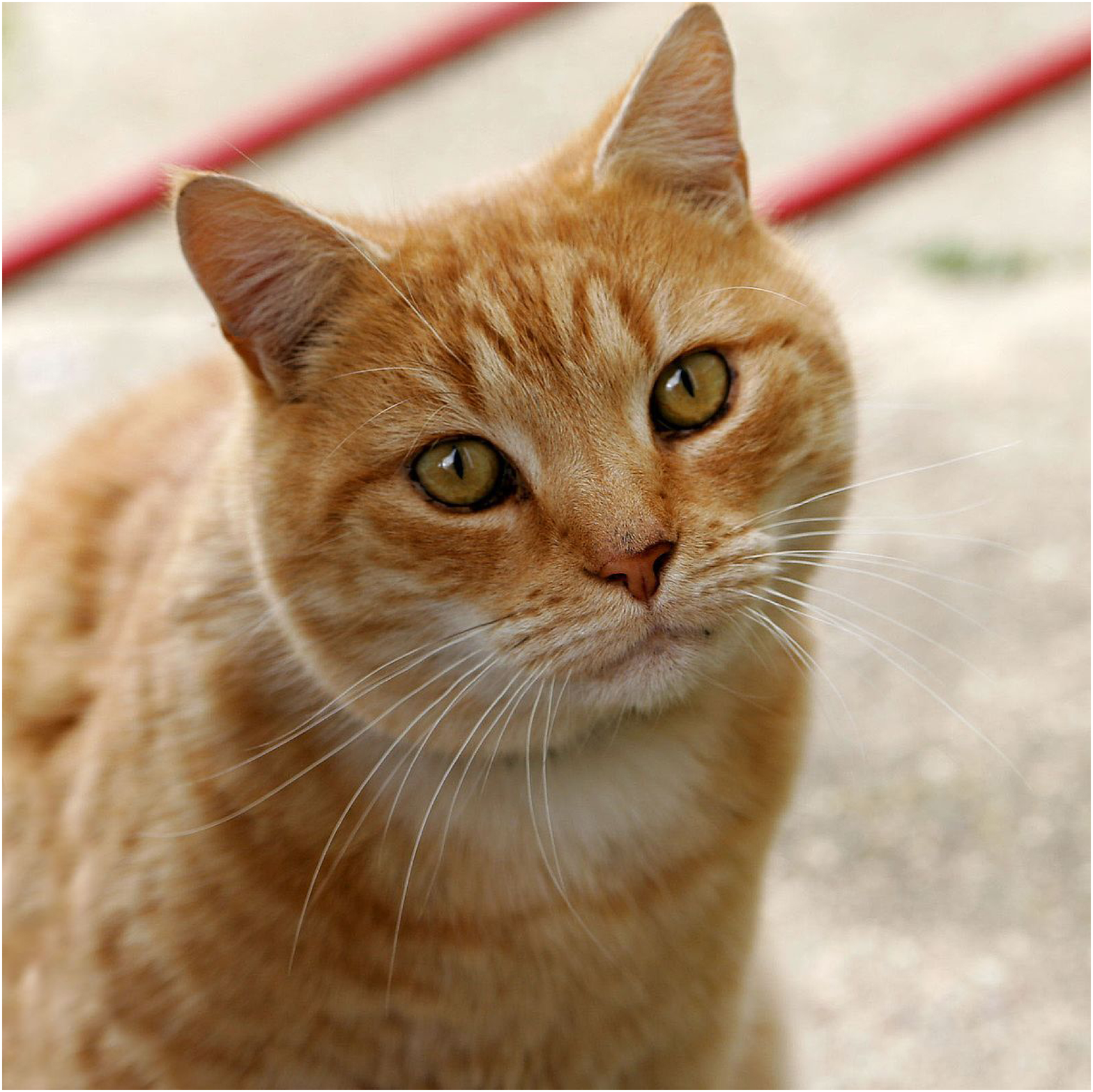}}{)}}\right \}$ and a set of features $T=\left\{digit, object\right\}$. Let us also assume a set of feature predictions $f = \left\{ \langle 2, digit, \emptyset \rangle, \langle cat, object, \emptyset \rangle \right \}$ and a label $l$ that corresponds to the set of unstructured raw data $D$. The context generator would then construct from the predictions $f$ the context $e_{ctx}= \{ \texttt{digit(2). object(cat).}\}$. 
\end{myexample}


\begin{mydef}[NSL example]
Let $(ID,X,l)$ be an input tuple where $ID$ is a unique identifier, $X$ is a set of samples of unstructured raw data with associated set $f_{X}$ of predicted features and a general aggregated confidence score $A_{X}$. Finally, let $l$ be a label from a given set $\mathcal{L}$ of possible labels. An NSL example $e$ is a \ac{wcdpi} given by the tuple $\left \langle ID, A_{X}, e_{pi}, \mathcal{G}(f_{X}) \right \rangle$, where the partial interpretation $e_{pi}$ is given by 
$\langle \{l\},\{l^{'}\mid l^{'}\in\mathcal{L}\setminus\{l\}\}\rangle$.
\end{mydef}

Algorithm~\ref{alg:nsl_example_construction} generates the NSL examples from a given labelled sample of unstructured raw data. 
\begin{algorithm}
\caption{NSL Example Generator}
\label{alg:nsl_example_construction}
\begin{algorithmic}[1]
\Procedure{generate}{$ID, X, l$}
    \State $INC = \left \{ l \right \};$
    \State $EXC = \{l^{'}\mid l^{'}\in\mathcal{L}\setminus\{l\}\};$
    \State $PRDCTS = [];$ $CONFS = [];$
    \For{$\langle d_{i}, t_{i}\rangle \in X$}
        \State $\rho_{d_{i}} = {n_{t_{i}}}(d_{i});$
        \State $y^{t_{i}}_{pred} = argmax(\rho_{d_{i}});$
        \State $y^{t_{i}}_{conf} = max(\rho_{d_{i}});$ 
        \State $PRDCTS \pluseq (y^{t_{i}}_{pred}, t_{i});$
        \State $CONFS \pluseq y^{t_{i}}_{conf};$
    \EndFor
    \State $CTX = \mathcal{G}(PRDCTS);$
    \State $PEN = \mathcal{A}(CONFS);$
    \State $WCDPI = (ID, PEN, (INC, EXC), CTX);$
    
    \State \textbf{return} $WCDPI;$ 
\EndProcedure
\end{algorithmic}
\end{algorithm}

Referring to Figure~\ref{fig:nsl_arch}, the identifier $ID$ for each NSL example is shown following the \texttt{\#pos} statement and the NSL example penalty $PEN$ calculated on line 13 in Algorithm~\ref{alg:nsl_example_construction} is shown in bold, immediately following the identifier. The NSL example penalty represents the aggregated confidence from neural network feature extractors. This encourages FastLAS to bias the learning towards hypotheses that cover NSL examples with high confidence neural network predictions, by ensuring that a high penalty is paid for leaving such examples uncovered by the learned hypothesis. The \textit{inclusion} and \textit{exclusion} sets $(INC, EXC)$ respectively, are shown in italics\footnote{For binary classification tasks such as the Sudoku board classification task shown in Figure~\ref{fig:nsl_arch}, we can simplify the inclusion and exclusion sets to only contain one label to represent the type of rules being learned. In this case, rules for \textit{invalid} Sudoku boards are learned, so \textit{valid} can be removed from the inclusion or exclusion set, depending on the example label.} in Figure~\ref{fig:nsl_arch}, followed by the \textit{context} $CTX$ given by the context generator $\mathcal{G}$.

\subsection{NSL Optimisation Criteria}
Our NSL framework uses the FastLAS system~\cite{law2020fastlas} for learning a hypothesis. FastLAS supports user-defined scoring functions that are capable of biasing the search for a hypothesis that is optimal with respect to a given domain-specific optimisation criteria. Also, FastLAS supports learning from noisy examples, where the notion of noise can be domain-specific. NSL makes use of these two features of FastLAS to maximise example coverage whilst minimising the hypothesis length, such that learned rules are interpretable and generalise over unseen examples, whilst taking into account the confidence of the neural network predictions. Specifically, NSL considers the noise of an example to be the penalty value given by the general aggregation score of the example's context predicted by the neural network, and defines a scoring function that combines the standard ILP optimisation criteria, for minimising rule length to encourage generalisation, with an example penalty that caters for the confidence estimation of the neural network predictions. 

\begin{mydef}[Penalty scoring function]\label{def:example_noise_sf}
Let $H$ be a hypothesis and $E$ be a set of generated NSL examples. A penalty scoring function $\mathcal{S}_{penalty}(H,E) = \sum\limits_{e \in UNCOV(H, E)}e_{pen}$ where $e_{pen}$ is obtained from line 13 in algorithm~\ref{alg:nsl_example_construction}, and $UNCOV(H,E)$ is the set of examples in E that are \textit{not covered} by $H$. Note that $H$ is said to cover an example $e$ iff $H$ accepts $e$~\cite{law2020fastlas}. 
\end{mydef}
The standard ILP scoring function~\cite{muggleton1994inductive} is given by the {\em length scoring function} $\mathcal{S}_{len}(H) = \left | H \right |$ which scores a hypothesis $H$ simply by counting the number of literals in $H$.

As FastLAS supports user-defined scoring functions, in our NSL framework, we can define a scoring function that combines the standard length scoring function $\mathcal{S}_{len}$ with the penalty scoring function given in Definition~\ref{def:example_noise_sf}, to jointly achieve both optimisation objectives. This is formally defined as follows: 

\begin{mydef}[NSL scoring function]\label{def:nsl_sf}
Let $H$ be a set of rules and $E$ be a set of generated NSL examples. The NSL scoring function $\mathcal{S}_{NSL}(H,E) = \mathcal{S}_{penalty}(H,E) + \gamma \mathcal{S}_{len}(H)$ where $\gamma \in [0, \infty]$ can be used to weight example coverage or generalisation. In this paper, we assume $\gamma = 1$.
\end{mydef}

We can now define the notion of an NSL learning task. Informally, this is based on the notion of an Observational Predicate Learning task ($ILP^{OPL}_{LAS}$) under the Answer Set semantics~\cite{law2020fastlas}, where examples are labelled unstructured data.

\begin{mydef}[NSL learning task]\label{def:nsl_learning_task}
An {\em NSL learning task} is a tuple $NSL_{task} = \langle  B, S_M, \mathcal{S}_{NSL}, E^{RAW}\rangle$ where $B$ is an \ac{asp} program called background knowledge, $S_M$ is the set  of possible rules, whose head predicates are the given labels and  body predicates are predicates in $B$ or in the given set of features $T$, $\mathcal{S}_{NSL}$ is the NSL scoring function and $E^{RAW}$ is a set of labelled raw unstructured data. 
\end{mydef}

The goal of the NSL learning task is to find an {\em optimal solution} to the task where the notion of {\em optimality} is given below.

\begin{mydef}[NSL Optimal Solution]
Let $LT=\langle  B, S_M, \mathcal{S}_{NSL}, E^{RAW}\rangle$ be an $NSL_{task}$ learning task. 
A set of rules $H\in S_{M}$ is an {\em optimal solution} of $LT$ if and only if there is no $H^{\prime} \in S_{M}$ such that $\mathcal{S}_{NSL}(H^{\prime},E) < \mathcal{S}_{NSL}(H,E)$ where $E$ is the set of NSL examples generated from the given $E^{RAW}$.
\end{mydef}


\section{Evaluation}
\label{evaluation}
To evaluate our NSL framework, we firstly focus on the ability to learn generalised and interpretable rules in the presence of perturbed training data. Secondly, we evaluate predictive performance at run-time where test data is perturbed proportionally to the amount of perturbations present at training time. 

Our evaluation is constructed over two neural-symbolic learning tasks: \begin{enumerate}
  \item \textbf{Zoo animal multi-class classification} where the goal is to classify an animal as either a \textit{mammal}, \textit{bird}, \textit{fish}, \textit{reptile}, \textit{bug}, \textit{amphibian}, or \textit{invertebrate} from a variety of features such as the number of legs. For this task, we use the `Zoo' data set\footnote{\label{zoo_url}\url{https://archive.ics.uci.edu/ml/datasets/Zoo}} which contains 101 examples. We select 10 numeric features and substitute digit feature values for an image of a corresponding digit from the MNIST test set. This task can be solved using propositional rules.
  
  \item \textbf{Sudoku board binary classification} as illustrated in Figure~\ref{fig:nsl_arch} where the goal is to learn rules that classify a 4x4 Sudoku board as \textit{valid} or \textit{invalid}. Hanssen's Sudoku puzzle generator\footnote{\label{sudoku_url}\url{https://www.menneske.no/sudoku/2/}} is used to generate 200 valid boards. We generate a further 200 invalid boards manually. In this task, digits on the Sudoku board are also replaced with images of corresponding MNIST test set digits. This task requires first-order rules (i.e. rules with variables) to correctly differentiate between valid and invalid Sudoku boards.
\end{enumerate}
We perform 5-fold cross-validation in both learning tasks with an 80\%/20\% train/test split. Full details regarding the data sets are listed in Section~\ref{sec:datasets}.

For the zoo animal classification task we compare our NSL framework to a shallow feed forward neural network and random forest baselines\footnote{Details of selected hyper-parameters and model architectures are given in Section~\ref{sec:hyper_params}}, as neural networks and tree-based methods are known to perform well on this task~\cite{nasser2019artificial, rfstudy}. Also, their learned models can be evaluated for interpretability by either using a surrogate model to approximate the predictions of a neural network, or inspecting learned trees directly in the case of a random forest.

For the Sudoku board classification task we compare NSL to a CNN-LSTM deep neural network architecture due to the spatial dependency between the sequence of digits on a Sudoku board. This architecture has demonstrated strong performance on other sequence classification tasks~\cite{8249013, zhou2015c}. We also compare NSL to a random forest model due to its inherently explainable properties.

Our evaluation uses two neural network feature extractors pre-trained on the MNIST training set. Their weights remain fixed throughout our experiments. We use a LeNet 5 architecture~\cite{lecun1998gradient} with the standard softmax classification layer and also a state-of-the-art \textit{uncertainty-aware} network called \textit{EDL-GEN}~\cite{sensoy2020uncertainty}\footnote{We use the softmax and EDL-GEN feature extractor implementation from \url{https://muratsensoy.github.io/uncertainty.html} and \url{https://muratsensoy.github.io/gen.html} respectively.}. To ensure a fair comparison with our NSL framework, in both learning tasks, all neural network and random forest baselines are given softmax feature extractor predictions as input and don't learn over the raw unstructured data directly. When aggregating softmax or EDL-GEN confidence information within the construction of an NSL example, we use the Gödel t-norm aggregation function due to the conjunctive nature of the learning tasks presented in this paper. The classification label in both tasks depends on the conjunction of animal features, or the conjunction of digits on a Sudoku board.

In both learning tasks, we simulate distributional shift for the neural network feature extractors by perturbing all the unstructured raw data within an increasing percentage of training examples. Unstructured raw data is perturbed by rotating each MNIST image within an example 90$^{\circ}$ clockwise~\cite{sensoy2018evidential}\footnote{Note. This is one example of a distributional shift.}. The final labels (i.e. the animal classification or the Sudoku board classification) remain unchanged.

\subsection{Learning rules from perturbed training data}
\begin{figure*}[!htb]
\centering
\begin{subfigure}{.24\textwidth}
  \centering
  \includegraphics[width=0.95\linewidth]{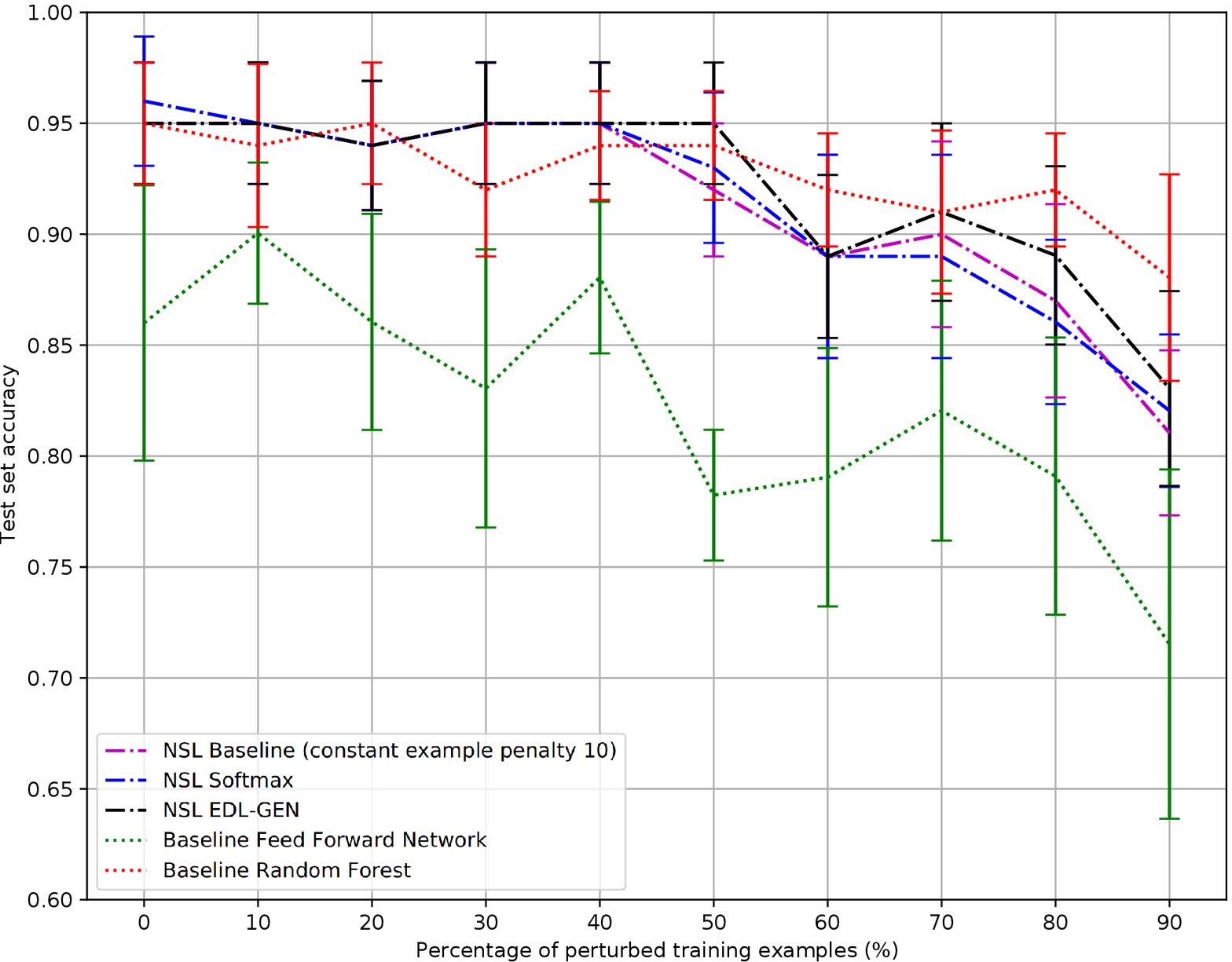}
  \caption{Zoo - accuracy}
  \label{fig:zoo_clean_test_data}
\end{subfigure}%
\begin{subfigure}{.24\textwidth}
  \centering
  \includegraphics[width=0.95\linewidth]{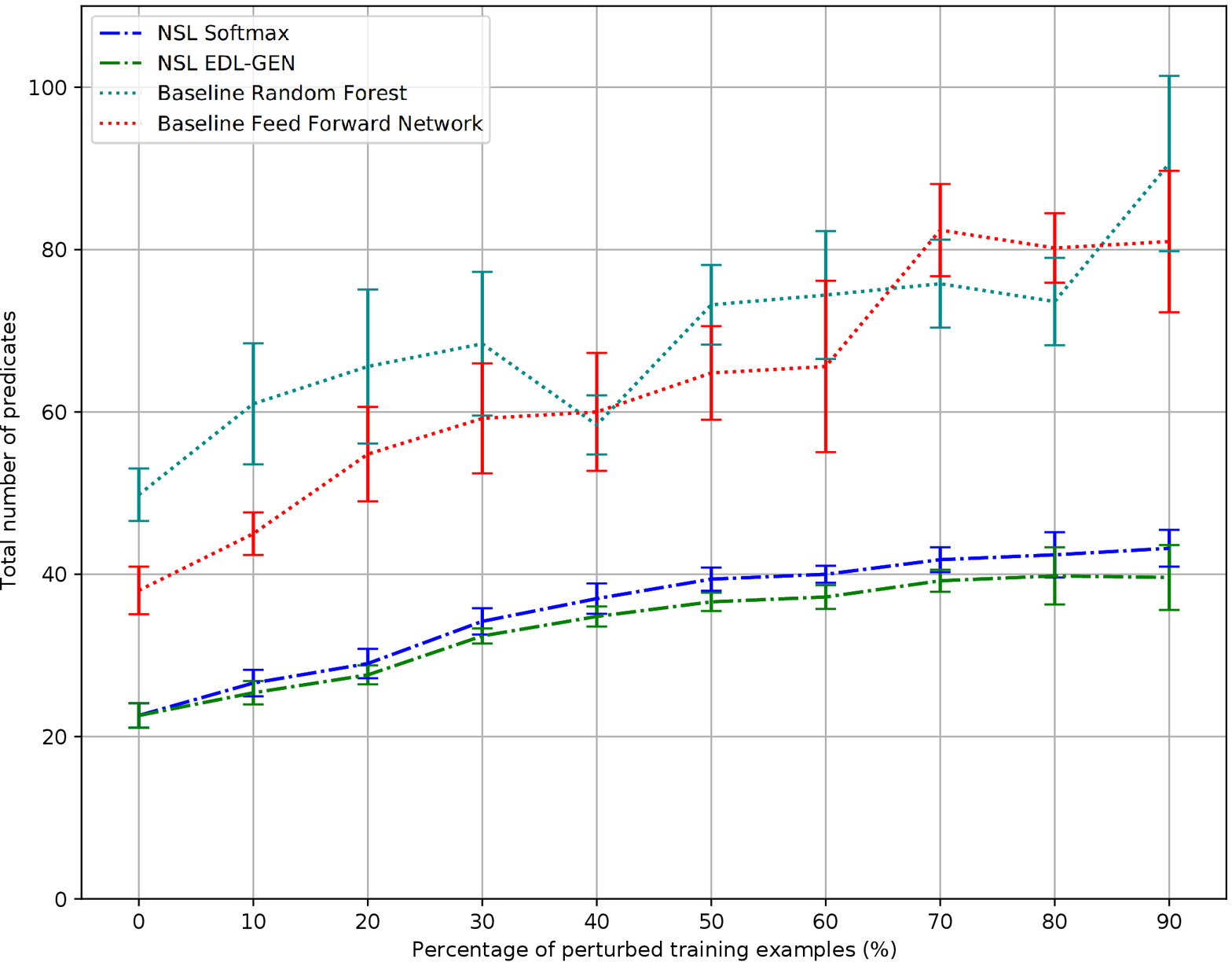}
  \caption{Zoo - Interpretability}
  \label{fig:zoo_interpretability}
\end{subfigure}
\begin{subfigure}{.24\textwidth}
  \centering
  \includegraphics[width=0.95\linewidth]{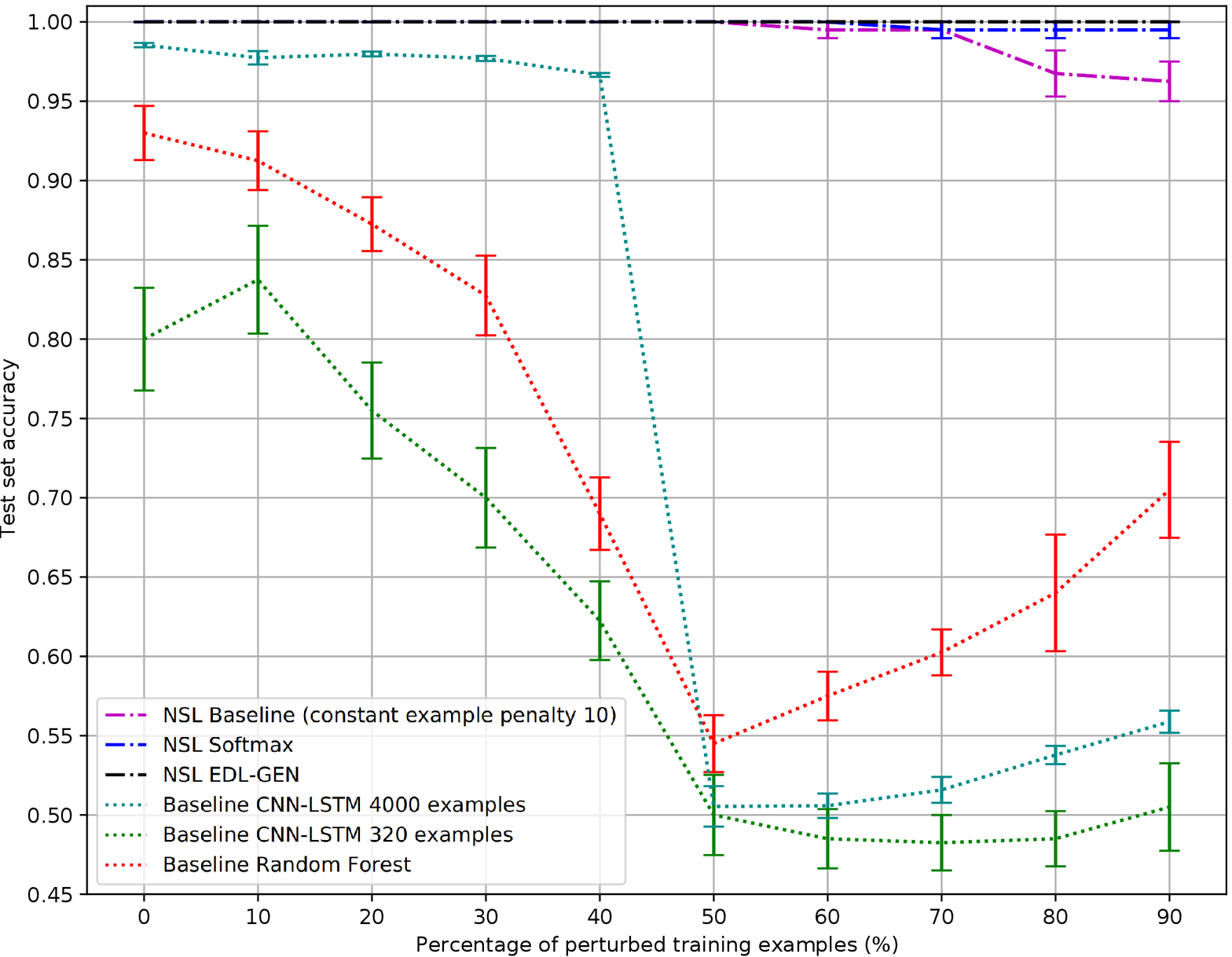}
  \caption{Sudoku - accuracy}
  \label{fig:sudoku_clean_test_data}
\end{subfigure}%
\begin{subfigure}{.24\textwidth}
  \centering
  \includegraphics[width=0.95\linewidth]{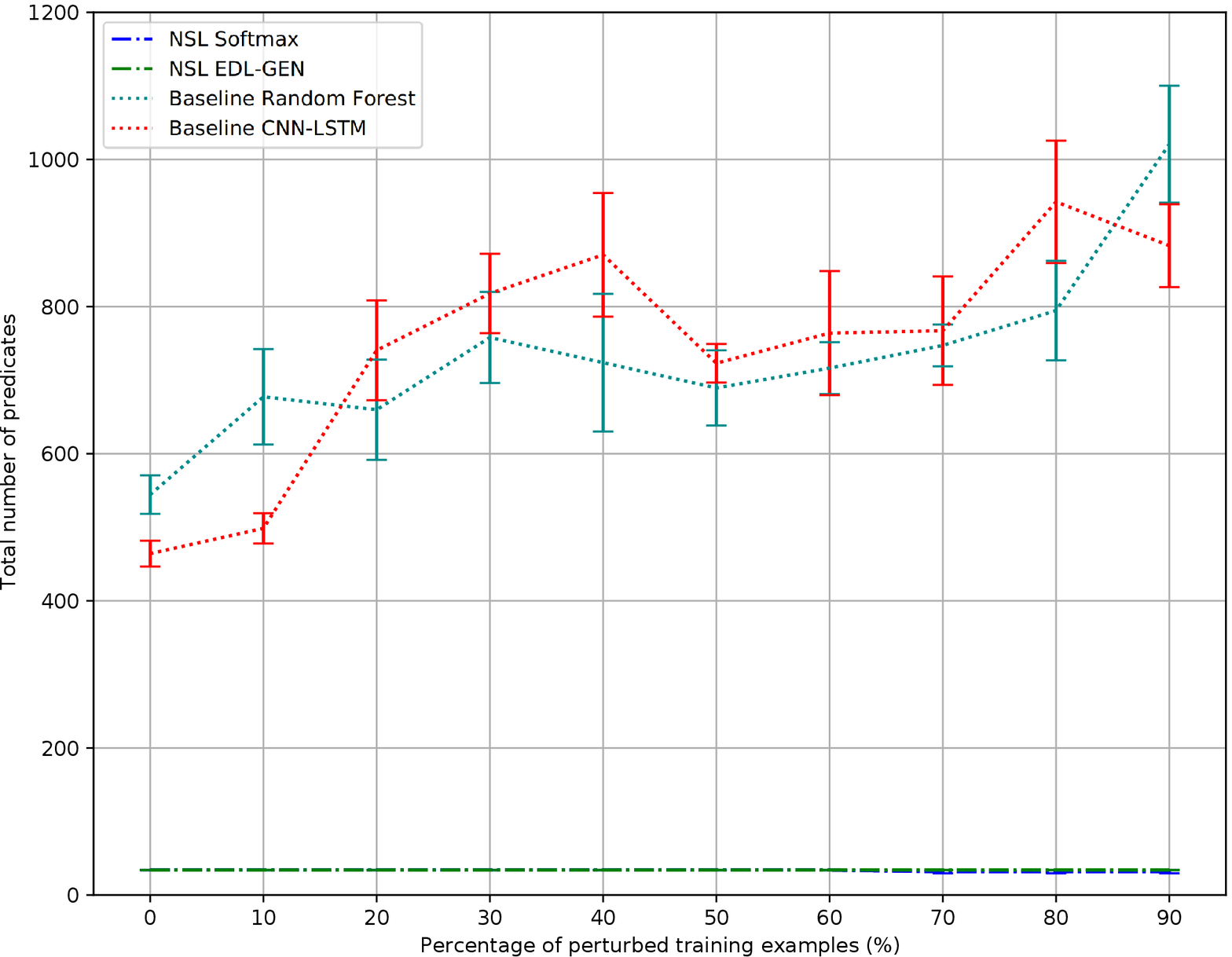}
  \caption{Sudoku - interpretability}
  \label{fig:sudoku_interpretability}
\end{subfigure}
\caption{Accuracy of learned rules with structured ground truth test data and interpretability of learned rules for both the zoo animal classification and Sudoku board classification tasks}
\label{fig:acc_interpretability_results}
\end{figure*}

Figures~\ref{fig:zoo_clean_test_data} and~\ref{fig:sudoku_clean_test_data} show the results for learning rules from perturbed training data on the zoo and Sudoku tasks respectively. In both tasks, the evaluation is performed on ground truth test data where the original feature values are used and are \textit{not} substituted for their corresponding MNIST digits. This is to indicate the quality of the learned rules and to evaluate the ability to ignore data perturbations at training time. The mean classification accuracy across all 5 cross-validation splits is reported to highlight classification performance. The standard error of the mean across all cross-validation splits is shown with the displayed error bars to indicate the deviation between the sampled mean and the true population mean. The \textit{NSL Baseline} uses a constant example penalty of 10 within FastLAS instead of aggregating neural network feature extractor confidence information. 


In terms of ground truth test set classification performance, NSL out-performs the baseline methods in both the zoo and Sudoku learning tasks, with the exception of the random forest in Figure~\ref{fig:zoo_clean_test_data}. Aggregating confidence information from neural network feature extractors leads to a benefit in both learning tasks after $\sim$40\% perturbations, with the EDL-GEN feature extractor outperforming softmax.

The zoo task in Figure~\ref{fig:zoo_clean_test_data} is relatively simple for symbolic \ac{ilp} learners and the neural network baseline even in a simple task is vulnerable to data perturbations, as NSL clearly demonstrates superior performance. NSL is marginally beaten by the random forest at 20\% and beyond 50\% perturbations, although NSL is within the standard error of the random forest. In order to obtain this increase in performance, the random forest learns a less interpretable model. To evaluate interpretability, we obtain the total number of predicates across all learned rules and plot the mean across all 5 cross-validation splits. The standard error of the mean across all splits is indicated with error bars. Rules with a lower number of predicates are assumed to be more interpretable~\cite{lakkaraju2016interpretable}. Figures~\ref{fig:zoo_interpretability} and~\ref{fig:sudoku_interpretability} show the total number of learned predicates increases as more perturbations are present in the training examples. In both the zoo and Sudoku learning tasks, NSL learns a more interpretable set of rules than the baseline methods for all percentages of training example perturbation and NSL with an EDL-GEN feature extractor outperforms NSL with a softmax feature extractor. 

We note that explaining what the random forest has learned may not be trivial, depending on the number of trees in the forest. In our experiments we have 100 trees in the ensemble. To interpret the neural network models, a surrogate decision tree is trained to approximate the neural network predictions~\cite{molnar2020interpretable}. For the random forest models, rules learned from the first tree within the ensemble are selected.

In Figure~\ref{fig:sudoku_clean_test_data}, which is a more complicated task requiring the use of memory, NSL outperforms all baselines for all percentages of training example perturbations. NSL is able to take advantage of background knowledge regarding the layout of a Sudoku board to increase its performance. At 50\% perturbations the performance of all the neural network and random forest baselines reduce to $\sim$50\% and start to increase again as more patterns begin to emerge in the training data. Given the CNN-LSTM is a deep neural network architecture, we conducted a further experiment with an increased number of training examples. We found that at 4000 training examples the CNN-LSTM was able to achieve similar performance to NSL with 320 training examples below 50\% example perturbations. This is a 12.5X reduction, indicating the data efficiency of the NSL framework as a result of background knowledge integration within FastLAS.


\subsection{Run-time perturbations}
\begin{figure*}[!htb]
\centering
\begin{subfigure}{.24\textwidth}
  \centering
  \includegraphics[width=0.95\linewidth]{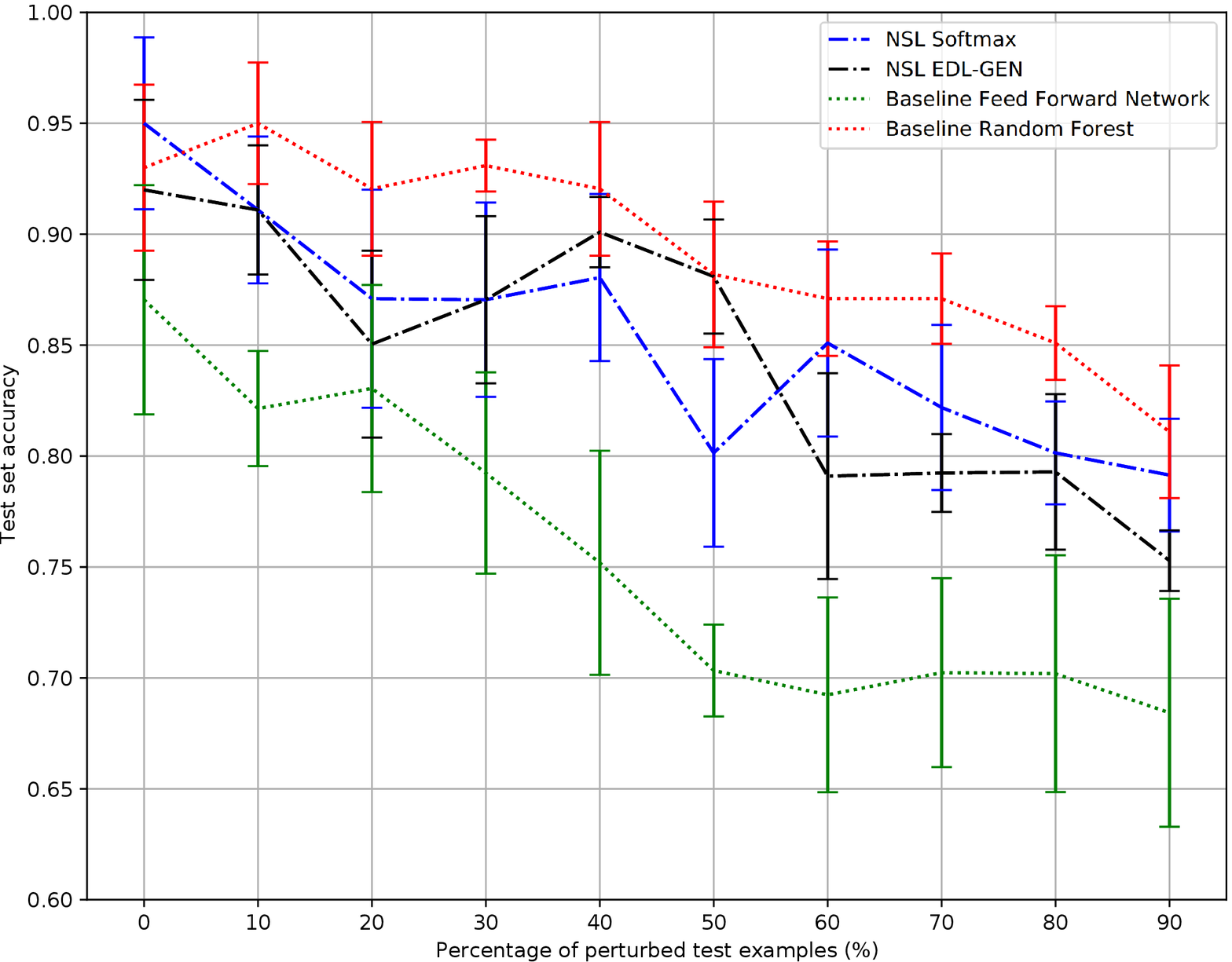}
  \caption{Zoo - accuracy}
  \label{fig:zoo_perturbed_acc}
\end{subfigure}%
\begin{subfigure}{.24\textwidth}
  \centering
  \includegraphics[width=0.95\linewidth]{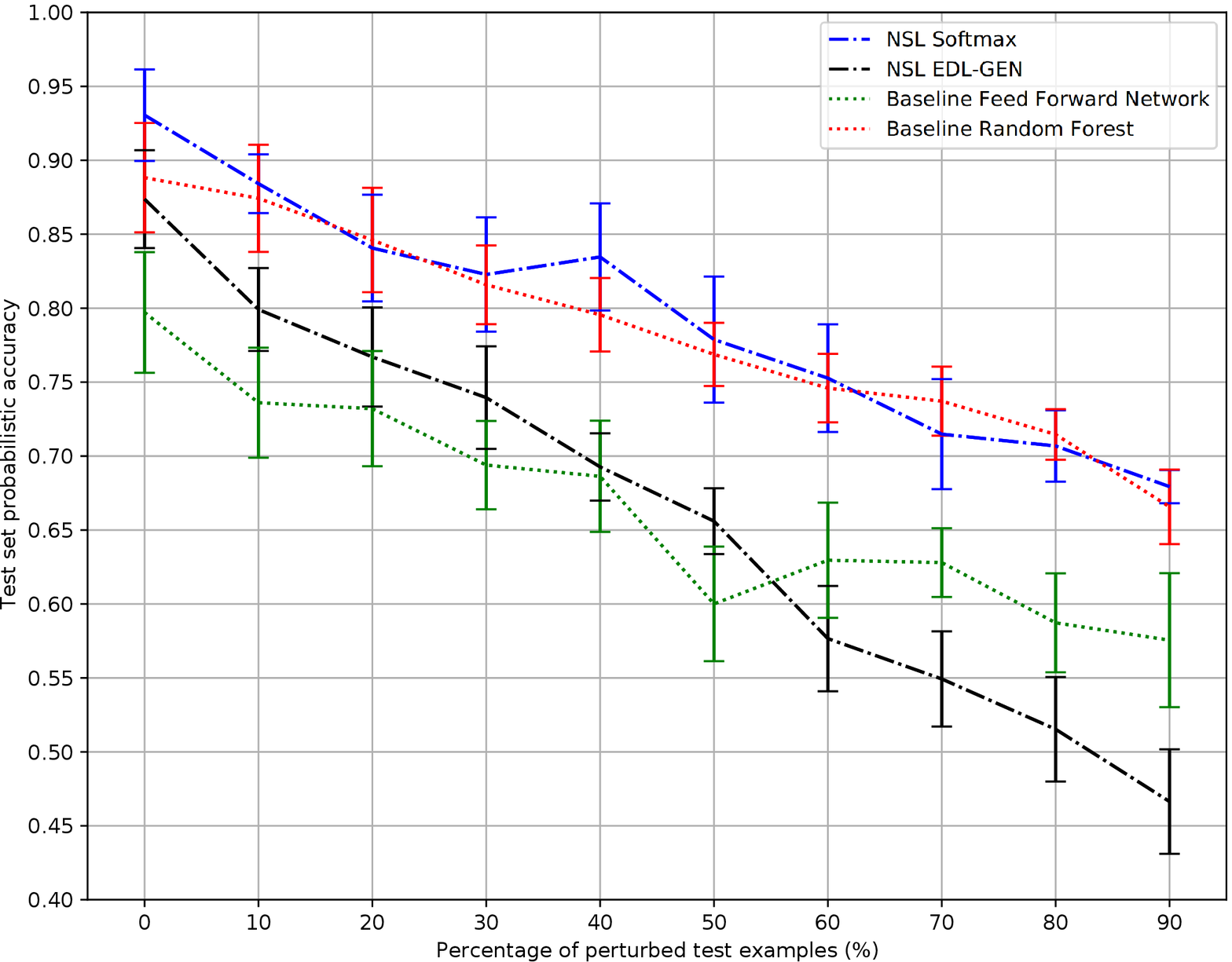}
  \caption{Zoo - prob. accuracy}
  \label{fig:zoo_perturbed_prob_ac}
\end{subfigure}
\begin{subfigure}{.24\textwidth}
  \centering
  \includegraphics[width=0.95\linewidth]{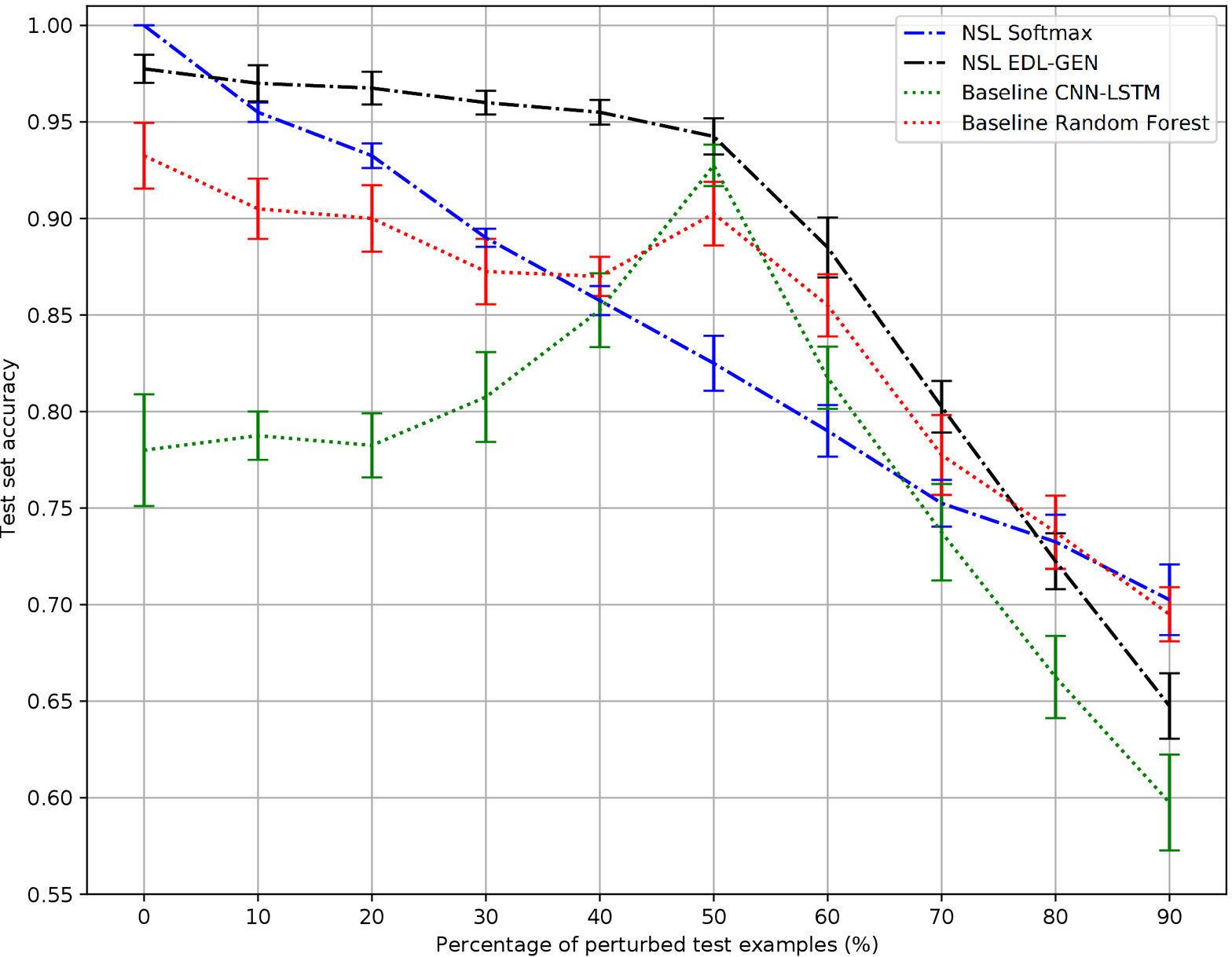}
  \caption{Sudoku - accuracy}
  \label{fig:sudoku_perturbed_acc}
\end{subfigure}%
\begin{subfigure}{.24\textwidth}
  \centering
  \includegraphics[width=0.95\linewidth]{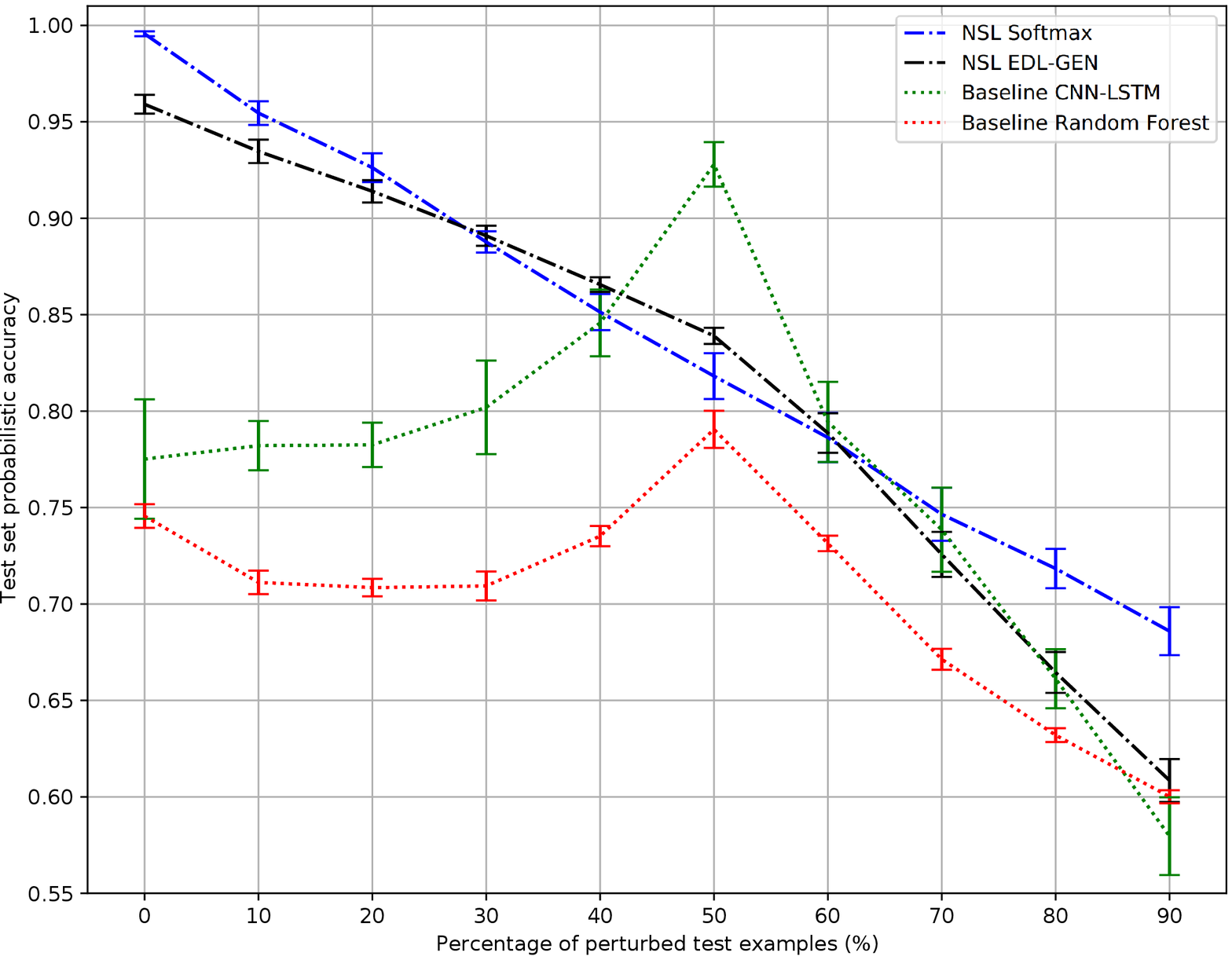}
  \caption{Sudoku - prob. accuracy}
  \label{fig:sudoku_perturbed_prob_acc}
\end{subfigure}
\caption{Run-time perturbation results for both the Zoo and Sudoku learning tasks, evaluated with accuracy and prob. accuracy}
\label{fig:run_time_peturb_results}
\end{figure*}

This subsection presents results where perturbations are applied to unstructured data at run-time in the same manner and at the same proportion that was observed during training in Figure~\ref{fig:run_time_peturb_results}, for both the zoo and Sudoku learning tasks.

In order to account for unconfident neural network predictions at run-time, we convert the rules learned with NSL to a ProbLog program~\cite{problog} and create annotated disjunctions~\cite{shterionov2015most} to represent the probability of a given MNIST image being classified as a certain digit. ProbLog then outputs a prediction as well as a probability of the prediction being true with respect to the neural network feature extractions and the learned rules. This is interpreted as a confidence value to align with the baseline neural network and random forest methods which already contain built-in confidence outputs at run-time.

To evaluate both the prediction and the confidence simultaneously, in addition to the standard accuracy metric, we use a modified accuracy metric, denoted \textit{probabilistic accuracy}, or \textit{prob. accuracy} for short, calculated as $\hat{a} = \frac{\sum_{i=1}^{n} p^{i}_{y\_true}}{n}$ where $p^{i}_{y\_true}$ represents the predicated confidence for the \textit{ground-truth} class for the $i$th test example and $n$ is the number of test examples. Instead of simply counting a correct (resp. incorrect) prediction as 1 (resp. 0) as with standard accuracy, this ensures that the learner will be penalised for predicting correctly with low confidence and also predicting incorrectly with high confidence. Error bars indicate the standard error when evaluating learned models from all 5 cross-validation splits. For these experiments, the \textit{NSL Baseline} is removed given we are evaluating in the presence of perturbed test data and unconfident neural network predictions.


For the zoo experiments in Figure~\ref{fig:zoo_perturbed_acc}, both NSL approaches outperform a neural network baseline and when evaluating with probabilistic accuracy in Figure~\ref{fig:zoo_perturbed_prob_ac}, NSL softmax outperforms the random forest apart from 20\%, 60\%, 70\% and 80\% perturbations. In the more complex task of learning rules for invalid Sudoku boards, both NSL approaches outperform the baselines until 40\% perturbations in both Figure~\ref{fig:sudoku_perturbed_acc} and~\ref{fig:sudoku_perturbed_prob_acc}, with the NSL EDL-GEN outperforming all approaches until 80\% perturbations in Figure~\ref{fig:sudoku_perturbed_acc}. Comparing Figures~\ref{fig:sudoku_perturbed_acc} and~\ref{fig:sudoku_perturbed_prob_acc} with Figure~\ref{fig:sudoku_clean_test_data}, it's likely that at 50\% perturbations the neural network has over-fitted to the pattern of perturbations in the training data and hasn't learned the true rules, given its strong performance when test data is perturbed and low performance when the test data is not perturbed.


\subsection{Data set Preparation}\label{sec:datasets}
\begin{enumerate}
    \item \textbf{Zoo animal classification} This data set is obtained from the UCI Machine Learning repository and contains 101 examples. In our experiments, we select the \textit{hair}, \textit{feathers}, \textit{eggs}, \textit{milk}, \textit{aquatic}, \textit{predator}, \textit{fins}, \textit{legs} and \textit{tail} features to enable experiments to be completed in a timely manner. The goal is to classify an example as either a \textit{mammal}, \textit{bird}, \textit{fish}, \textit{reptile}, \textit{bug}, \textit{amphibian}, or \textit{invertebrate}. We perform 5-fold cross validation to generate 5 training sets each containing 80 examples and 5 test sets each containing 21 examples respectively. Digit features are then substituted for a random image from the MNIST test set, corresponding to each digit. This is initially performed for the 5 zoo training sets and for the run-time perturbation experiments this is performed for the 5 zoo test sets also.
    
    \item \textbf{Sudoku board classification} This data set is generated using 4x4 Sudoku puzzles obtained from Hanssen's Sudoku puzzle generator. We obtain 200 valid puzzles and generate a further 200 invalid puzzles manually to create a data set of 400 examples. The goal is to classify whether a Sudoku board is \textit{valid} or \textit{invalid} based on the digits present in the board. Each example contains a varying number of digits and no example details a completed board. Similarly to the zoo animal classification data set, we perform 5-fold cross validation to generate 5 training sets each containing 320 examples and 5 test sets each containing 80 examples respectively. When evaluating the CNN-LSTM architecture with a larger data set, we obtain 2500 valid Sudoku puzzles from Hanssen's Sudoku puzzle generator and generate a further 2500 invalid puzzles manually. 5-fold cross-validation is then performed to generate 5 training sets of 4000 examples and 5 test sets of 1000 examples respectively. We also perform a similar substitution of digits with random images from the MNIST test set corresponding to the same digit.
\end{enumerate}

\subsection{Baseline Architectures and Hyper-Parameter Tuning}\label{sec:hyper_params}
\begin{enumerate}
    \item \textbf{Zoo animal classification} The baseline feed forward network consists of 3 fully connected layers where the first two layers contain \textit{ReLU} activation functions and the final layer contains a \textit{softmax} activation function. The number of neurons in the first two layers as well as the batch size were tuned using a grid search with the following parameter values. Number of neurons in layer 1: $\{5,12,20\}$, number of neurons in layer 2: $\{8,10,15\}$ and batch size: $\{2,5,10\}$. Tuning was performed over the extended zoo dataset `\textit{zoo3}'\footnote{https://www.kaggle.com/agajorte/zoo-animals-extended-dataset} with no MNIST digit substitution where the best performing combination of parameters were chosen according to test set accuracy. We performed an 80\%/20\% train/test split on the zoo3 data set using the scikit-learn library. All random seeds were set to 0 and the chosen parameters were 12 neurons in the first layer, 8 neurons in the second layer and a batch size of 10. The model was trained for 150 epochs with the sparse categorical cross-entropy loss using the Adam optimiser and was implemented in TensorFlow 2.3.0 with Keras 2.4.0.
    
    The baseline random forest model was implemented with scikit-learn 0.23.2 and tuned on the same train/test split of the zoo3 dataset as the feed forward neural network. The number of estimators were tuned across: $\{10,50,100,200\}$. The best performing parameter value of 100 estimators was chosen. 
    
    \item \textbf{Sudoku board classification} The baseline CNN-LSTM network consists of an embedding layer, followed by a 1D convolutional layer with 32 filters, a kernel size of 3, the ReLU activation function and `same' padding. Then, a 1D max pooling layer with pool size 2 is used, followed by a dropout layer with dropout probability 0.2, an LSTM layer and a second dropout layer also with dropout probability 0.2. Finally, a dense fully connected layer with the sigmoid activation function is used to produce a binary classification of the input digit sequence. The input sequence length to the embedding layer is 16, representing each cell on the 4x4 Sudoku board.
    
    Tuning was performed using a separate dataset of 100 valid and 100 invalid Sudoku boards using a grid search to tune the embedding vector length, the number of neurons in the LSTM layer and the batch size with the following parameter values. Embedding vector length: $\{10,20,32\}$, number of neurons in the LSTM layer: $\{50,100\}$ and batch size: $\{8,16,32\}$. The best performing combination of parameter values was chosen according to test set accuracy. We performed an 80\%/20\% train/test split using the scikit-learn library. All random seeds were set to 0 and the chosen parameters were an embedding vector length of 32, 100 neurons in the LSTM layer and a batch size of 32. The model was trained for 200 epochs with the binary cross-entropy loss using the Adam optimiser and was implemented in TensorFlow 2.3.0 with Keras 2.4.0.
    
    The baseline random forest model was implemented with scikit-learn 0.23.2 and tuned on the same train/test split of the separate Sudoku dataset as the CNN-LSTM neural network. The number of estimators were tuned across: $\{10,50,100,200\}$. The best performing parameter value of 100 estimators was chosen. 
\end{enumerate}

\subsection{System Details}
All experiments in this paper were run on the same machine with the following specifications:

\begin{itemize}
    \item \textbf{Hardware:} QEMU KVM virtual machine standard PC (i440FX + PIIX 1996) with 10 nodes of 8-core AMD EPYC Zen 2 CPUs (80 cores total), 16GB RAM.
    \item \textbf{Operating System:} Ubuntu 18.04.4 LTS.
    \item \textbf{Software:} FastLAS 1.0, Python 3.7.3, TensorFlow 2.3.0, Keras 2.4.0, scikit-learn 0.23.2, numpy 1.19.1, problog 2.1.0.42. 
\end{itemize}

\section{Related Work}
Existing approached for integrating neural and symbolic techniques can be categorised into two main types. Firstly, there are a set of approaches that leverage neural networks for learning and symbolic components for reasoning. Secondly, there are approaches that leverage both neural networks and symbolic components together to perform a joint learning task.

DeepProbLog~\cite{manhaeve2018deepproblog} is an example of an integrated neural-symbolic learning and reasoning framework that connects probabilistic logic to neural networks to enable probabilistic reasoning over neural network outputs. Neural network feature extractors within DeepProbLog can be trained w.r.t resulting policy labels, by constructing a Sentential Decision Diagram (SDD) based on hard-coded logical rules. DeepProbLog requires these logical rules to be manually specified and they are not learned from data.


Logic Tensor Networks (LTNs)~\cite{serafini2016logic} generalise the semantics of first-order logic to introduce real logic, replacing the standard Boolean values with real values in the interval $[0,1]$. LTNs enable integrated learning through tensor networks and reasoning using real logic. This enables soft and hard logical constraints and relations to be specified in first-order logic as background knowledge to help guide neural networks during training. 
With LTNs, logical rules aren't learned symbolically, learning occurs in a neural network constrained by real logic. 

In both of these approaches, logical rules are not directly learned from unstructured data, which is the key focus of our NSL framework. A differentiable \ac{ilp} framework~\cite{evans2018learning} called $\partial ILP$ learns rules from unstructured data by re-implementing a top-down, generate and test \ac{ilp} approach. $\partial ILP$ learns, through stochastic gradient descent, which generated clauses should be ``turned on" such that together with the background knowledge, the turned on clauses entail the positive examples and do not entail the negative examples. Their evaluation includes a pre-trained neural network that classifies MNIST digits connected to an \ac{ilp} system for rule learning. Their approach requires, a set of predefined rule templates to define the hypothesis space. This is memory intensive and the approach is limited to predicates up to arity 2. NSL does not have any such constraints as FastLAS is more scalable than the \ac{ilp} system used in $\partial ILP$ in terms of the number of examples and the size of the hypothesis space. For example, $\partial ILP$ cannot be used on the Mutagenesis data set\footnote{https://relational.fit.cvut.cz/dataset/Mutagenesis}, which contains 230 examples, whereas the Sudoku task presented in this paper contains 320 training examples. Finally, the differentiable integration in $\partial ILP$ is tightly coupled with the \ac{ilp} system, whereas NSL is modular and preserves separation between the neural and symbolic components. 
\section{Conclusion}
This paper has introduced a hybrid neural-symbolic learning framework called \textit{NSL} that is capable of learning robust rules in the presence of perturbed training data and incorrect or unconfident neural network predictions. The framework aggregates confidence information from neural network feature extractors to create a penalty paid by the FastLAS \ac{ilp} system for not covering an NSL example. Our results indicate that NSL is able to learn robust rules in the presence of perturbed training data, matching or outperforming the neural network and random forest baselines whilst learning a more general and interpretable model. When perturbations occur at run-time, NSL is able to maintain robust classification performance up to $\sim$40\% perturbations, outperforming the neural network and random forest baselines. Finally, in the Sudoku task, FastLAS can utilise background knowledge to learn robust rules with fewer training examples compared to a neural network baseline.


\section{Ethics Statement}
Interpretable machine learning systems provide a first step towards addressing issues such as bias, discrimination and accountability. In practical terms, it often isn't possible to acquire training data that perfectly represents the intended distribution within a population, one can only approximate the distribution through a set of samples. Powerful machine learning algorithms, such as deep neural networks may unwillingly amplify and reinforce inherent bias present in the sampled underlying training data that can be difficult to detect until the model is deployed. 

In this paper, we explore the combination of an interpretable \ac{ilp} system with a deep neural network feature extractor such that learned rules from perturbed data can be inspected for correctness and potential bias or discrimination. An interesting example of the benefit of this approach can be observed in the Sudoku board classification task within our evaluation. Referring to Figures~\ref{fig:sudoku_perturbed_acc} and~\ref{fig:sudoku_perturbed_prob_acc} when evaluating on perturbed test data, the CNN-LSTM neural network baseline appears to perform strongly, given 50\% perturbed training and test examples. However, comparing the performance of the same model when evaluated on \textit{non-perturbed} test data in Figure~\ref{fig:sudoku_clean_test_data}, the performance is significantly reduced. This indicates the CNN-LSTM has amplified the perturbations present in the training data. This particular model may be difficult to interpret which highlights a potential issue should the training data contain biased or discriminatory examples. An interpretable approach, such as the NSL framework presented in this paper, could highlight such amplifications of biased training data before the model is deployed.

\section{Acknowledgments}
This research was sponsored by the U.S. Army Research
Laboratory and the U.K. Ministry of Defence under Agreement
Number W911NF-16-3-0001. The views and conclusions
contained in this document are those of the authors and should
not be interpreted as representing the official policies, either
expressed or implied, of the U.S. Army Research Laboratory,
the U.S. Government, the U.K. Ministry of Defence or the
U.K. Government. The U.S. and U.K. Governments are
authorized to reproduce and distribute reprints for Government
purposes notwithstanding any copyright notation hereon.

\bibliographystyle{unsrt}  
\bibliography{references}  

\end{document}